\documentclass[10pt,twocolumn,letterpaper]{article}

\usepackage{wacv}              

\definecolor{wacvblue}{rgb}{0.21,0.49,0.74}
\usepackage[pagebackref,breaklinks,colorlinks,allcolors=wacvblue]{hyperref}

\DeclareRobustCommand{\cuneiformA}{%
  \raisebox{-0.25\height}{%
    \includegraphics[height=1.8ex]{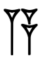}%
  }%
}
\usepackage[pagebackref,breaklinks,colorlinks]{hyperref}
\usepackage{comment}
\usepackage[accsupp]{axessibility}  
\usepackage{float}
\usepackage[capitalize]{cleveref}
\crefname{section}{Sec.}{Secs.}
\Crefname{section}{Section}{Sections}
\Crefname{table}{Table}{Tables}
\crefname{table}{Tab.}{Tabs.}

\def\wacvPaperID{VisionDocs-16} 
\def\confName{WACV}
\def\confYear{2027}

\title{Domain-Specific Self-Supervised Learning for Cuneiform Sign Representation}

\author{%
  Asim Niaz$^{1,2}$ \quad Morris Alper$^{3}$ \quad Shai Gordin$^{4}$ \quad Enrique Jim\'enez$^{1}$\\[6pt]
  $^{1}${\small Ludwig-Maximilians-Universität München} \quad 
  $^{2}${\small Munich Center for Machine Learning (MCML)}\\[4pt]
  $^{3}${\small University of Miami} \quad 
  $^{4}${\small Ariel University}
}

\begin{document}
\maketitle
\begin{abstract}
Recognizing cuneiform signs is challenging: sign shapes vary across historical periods, scribal practices, and preservation conditions, while expert annotations are scarce. We study whether domain-specific self-supervised pretraining yields better cuneiform sign representations than generic natural-image pretraining. We pretrain a masked autoencoder, CuneiMAE, on over 900K unlabeled sign crops and compare it with ImageNet-pretrained MAE, CLIP, and DINOv2 on a 218-class sign recognition benchmark. CuneiMAE learns substantially stronger frozen representations ({43.7\% linear-probe accuracy vs.\ 25.2\% for ImageNet-MAE}), with
 {gains over ImageNet-MAE on cross-period transfer}. Under fine-tuning, CuneiMAE outperforms ImageNet-pretrained baselines on our internal benchmark, with the largest gains when labels are scarce ({+20.4 pp} over ImageNet-MAE with 1\% of labels).
 On an external benchmark,
{CuneiMAE reaches 69.9\% top-1 accuracy after fine-tuning, about 8 pp above the ResNet-18 baseline from prior work.}
Explicit structural priors, such as edge-guided masking and part bottlenecks, do not improve over standard masked reconstruction.
Overall, domain-specific self-supervision {shows strong performance on cuneiform recognition tasks, particularly when labels are limited.}
\end{abstract}    
\section{Introduction}
\label{sec:intro}

Cuneiform is one of the earliest documented writing systems in human history, preserving over three millennia of historical, administrative, and literary records across more than half a million discovered clay tablets. Automated historical document analysis of these artifacts could accelerate transcription, paleographic comparison, and dating. However, visual recognition remains challenging because individual signs exhibit substantial variation in wedge morphology, orientation, and ductus across centuries, while surface erosion and tablet damage introduce additional physical degradation \cite{bogacz2022digital}. These challenges are compounded by the limited availability of consistently annotated sign-level data, as reliable transcription and annotation require specialized expertise in Assyriology \cite{dencker2020deep}.

Self-supervised visual representation learning offers a promising route toward data-efficient historical document analysis by enabling encoders to learn useful visual representations from unlabeled imagery prior to downstream supervision~\cite{he2022masked}. This is particularly attractive for cuneiform, where unlabeled sign images are considerably easier to obtain than expert annotations. However, widely used pretrained visual encoders are typically learned from natural-image datasets such as ImageNet, whose visual statistics differ substantially from impressed wedge patterns on clay, and computational research on cuneiform has largely relied on supervised training or such generic backbones~\cite{stotzner2023cnn,rest2022illumination,yugay2024stylistic,williams2025deepscribe}. It therefore remains unclear whether domain-specific self-supervision yields representations better matched to the structural and stylistic variability of cuneiform signs.

We ask: Does domain-specific self-supervision improve the quality of cuneiform representations? How do such advantages depend on the amount of labeled data? Do the learned features transfer across distinct chronological periods and scribal traditions? {And do explicit structural priors, such as edge-guided objectives, improve on standard masked reconstruction?} These questions are particularly relevant in cultural-heritage settings, where expert labels are limited and target artifacts may differ substantially from the training distribution. 

To address these questions, we pretrain a Masked Autoencoder (MAE) with a Vision Transformer backbone (ViT-B/16) from scratch on a curated corpus of {940,602} unlabeled cuneiform sign crops, which we call CuneiMAE. Masked image modeling provides a natural self-supervised objective for cuneiform imagery, requiring the encoder to model relationships among partially observed wedge structures. This mirrors epigraphic practice, in which damaged signs are identified from their surviving wedges.

We evaluate the resulting representations against generic pretrained encoders (ImageNet-MAE, CLIP, and DINOv2) on a 218-class sign recognition benchmark, {and on an external CSDD-derived benchmark~\cite{dencker2020deep}, where we also test a checkpoint initialized from ImageNet-MAE and ProtoSnap's synthetic training data~\cite{mikulinsky2025protosnap}.}
The contributions of this work are threefold.

\begin{itemize}

    \item \textbf{Large-Scale Domain-Specific Pretraining:}
    {We show that MAE pretraining on cuneiform sign crops improves frozen linear probing by {18.45 pp} and k-NN accuracy by up to 8.85 pp over ImageNet-MAE, {and outperforms it in every cross-period transfer direction}.}

    \item \textbf{Label-Efficient and External Evaluation:}
    We show that domain-specific pretraining provides larger gains under limited supervision, with the advantage narrowing as more labeled data become available. On the external CSDD-derived benchmark, {both CuneiMAE checkpoints outperform ResNet-18 after fine-tuning without synthetic data, with the ImageNet-initialized checkpoint achieving the highest accuracy.}

    \item \textbf{Structure-Aware Pretraining Evaluation:}
    We evaluate edge-constrained masking, edge-weighted reconstruction, dual RGB-edge reconstruction, and a part-bottleneck variant against CuneiMAE, and find that none improves over standard masked reconstruction {under controlled 10-epoch pretraining}.

\end{itemize}

To facilitate reproducibility and encourage further work in digital epigraphy, {code}, pretrained checkpoints, and access instructions to the dataset {are} publicly available at \url{https://github.com/ElectronicBabylonianLiterature/cuneimae}.
\section{Related Work}

\subsection{Computer Vision for Cuneiform}
Machine learning has been applied to a range of ancient scripts~\cite{sommerschield2023machine,diao2025ancient,assael2019restoring,huang2019obc306}, with cuneiform posing particular challenges due to the visual complexity of impressed wedges on non-uniform, often damaged clay surfaces. Several works model sign structure explicitly, encoding wedge configurations as graphs~\cite{kriege2018recognizing} or localizing individual wedges~\cite{stotzner2023r,hamplova2024cuneiform,mikulinsky2025protosnap}. Other work applies deep models directly to photographs of tablets for sign detection and recognition~\cite{dencker2020deep,rusakov2020towards,williams2025deepscribe,cobanoglu2024sign,che2026automated}, including supervised sign classification~\cite{verwimp2025signs,yugay2024stylistic}, often supplemented with synthetic training data~\cite{rest2022illumination,stotzner2023cnn,brandenbusch2021context,mikulinsky2025protosnap}. We follow the latter approach in classifying sign crops from 2D photographs, but focus on the downstream effect of vision encoder pretraining.

Tablet and sign appearance varies widely across periods, scribes, and digitization conditions~\cite{kapon2024shaping,chen2024classification,yugay2024stylistic}, and models often transfer poorly across collections~\cite{verwimp2025signs,stotzner2023cnn,rest2022illumination}. We therefore also evaluate transfer across historical periods.

\subsection{Domain-Specific Self-Supervised Learning}
Self-supervised methods learn visual representations from unlabeled data using a variety of objectives, including contrastive~\cite{chen2020simple}, self-distillation~\cite{caron2021emerging,oquab2023dinov2}, and reconstruction-based~\cite{he2022masked} approaches; our MAE-based approach falls in the latter category. As foundation vision models are typically pretrained on natural images, self-supervised pretraining has proven effective for adapting to substantially different domains, such as medical imaging~\cite{zhou2023self}, remote sensing~\cite{cong2022satmae}, and, as in our case, visual text~\cite{yang2022reading,lyu2022maskocr,li2022dit}.

Self-supervision is especially attractive for historical documents, for which annotation is less readily available than for modern texts. {Related work on historical document layout analysis has compared in-domain and out-of-domain transfer~\cite{de2025domain}, while an overview reports benefits from document-specific pretraining over ImageNet pretraining and training from scratch~\cite{de2023imagenet}. Self-supervision} has improved low-resource historical transcription~\cite{vogler2022lacuna}, handwritten text recognition with limited labels~\cite{kivsvs2024self}, keyword spotting~\cite{jemni2026st}, writer identification~\cite{lastilla2022self}, and few-shot symbol classification~\cite{alfaro2023few}. For cuneiform, ProtoSnap~\cite{mikulinsky2025protosnap} uses fine-tuned diffusion features, but these are dense features used for local correspondence matching rather than global representations for recognition. To our knowledge, domain-specific self-supervised pretraining has not yet been systematically evaluated for cuneiform sign recognition.

\section{Method}
\label{sec:method}

Our study evaluates whether self-supervised pretraining on cuneiform imagery produces more useful visual representations than generic natural-image pretraining. We use masked autoencoding as the common self-supervised framework and compare an ImageNet-pretrained MAE with a ViT-B/16 MAE pretrained from scratch on unlabeled cuneiform sign crops. Using the same encoder architecture controls for differences in model capacity across the two pretraining regimes. {Only on the external CSDD-derived benchmark do we also evaluate CuneiMAE initialized from ImageNet-MAE and further pretrained on cuneiform.}

\begin{figure*}[t]
    \centering
    \includegraphics[width=\textwidth]{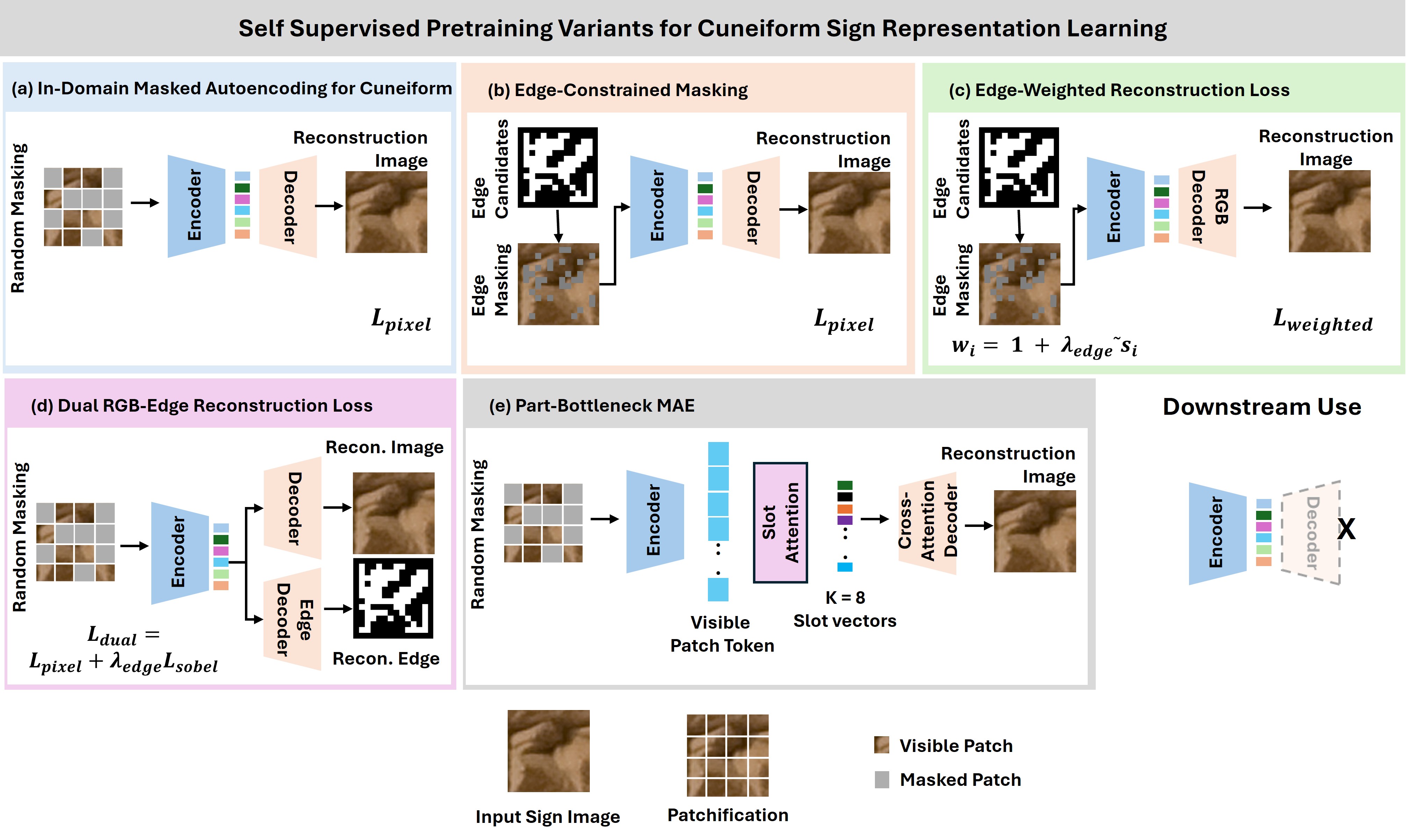}
    \caption{Overview of the self-supervised pretraining variants for cuneiform sign representation learning. During downstream evaluation, the MAE decoder is discarded and only the pretrained encoder is retained for feature extraction and downstream classification.}
    \label{fig:overview}
\end{figure*}

In addition, we evaluate several structure- and part-aware MAE variants. An overview of the pretraining designs and downstream evaluation is shown in Fig.~\ref{fig:overview}.

\subsection{In-Domain Masked Autoencoding for Cuneiform (CuneiMAE)}
Let $x \in \mathbb{R}^{H \times W \times C}$ denote an input cuneiform sign crop ($H=W=224, C=3$). The image is partitioned into non-overlapping patches of size $P \times P$ ($P=16$), yielding a sequence of $L = HW / P^2 = 196$ patch tokens. Patches are linearly projected into dimension $D = 768$ and summed with fixed 2D sine-cosine positional embeddings. Following the standard MAE framework, a fixed fraction $\rho = 0.75$ of patches is randomly masked without replacement. The unmasked visible tokens ($L_{\text{vis}} = (1 - \rho)L \approx 49$) are processed by a standard Vision Transformer encoder ($\text{ViT-B/16}$, 12 blocks, 12 attention heads):
\begin{equation}
    z_{\text{vis}} = \text{Encoder}(x_{\text{vis}})
\end{equation}
The latent tokens $z_{\text{vis}}$ are combined with shared, learnable mask tokens $[e_{\text{mask}}]$ to restore the full spatial sequence of length $L$. A lightweight Transformer decoder (8 blocks, dimension 512, 16 heads) processes the unmasked and mask tokens together with decoder positional embeddings to reconstruct the pixel values of the original patches:

\begin{equation}
\hat{x} = \text{Decoder}([z_{\text{vis}} \,\Vert{}\, e_{\text{mask}}]).
\end{equation}

The objective minimizes the mean squared error (MSE) evaluated strictly over the masked patch indices $\mathcal{M}$:
\begin{equation}
\mathcal{L}_{\text{pix}} = \frac{1}{\vert{}\mathcal{M}\vert{}} \sum_{i \in \mathcal{M}} \Vert{} x_i - \hat{x}_i \Vert{}_2^2 ,   
\end{equation}

where $x_i$ and $\hat{x}_i$ denote the normalized pixel targets and reconstructions for patch $i$.



\subsection{Structure- and Part-Guided Inductive Variants}
\label{sec:structure}

Sign identity is determined by wedge configuration, while much of each crop is uninformative clay surface. Following prior work that models wedge structure explicitly~\cite{kriege2018recognizing,stotzner2023r,hamplova2024cuneiform}, we test whether biasing pretraining toward edges or compact part-based representations improves the learned features. Each variant modifies a different MAE component and is evaluated as a separate ablation.

\subsubsection{Edge-Constrained Masking}

We use Sobel-gradient salience to prioritize masking of edge-rich patches. Edge40 masks only the sampled edge-prioritized patches, resulting in an effective masking ratio of 15.8\%. Edge75 supplements the same edge-prioritized set with randomly sampled patches to retain the standard 75\% masking ratio. Full salience computation and sampling details are provided in the supplementary material.

\subsubsection{Edge-Weighted Reconstruction Loss}

This variant retains standard 75\% random masking but assigns greater reconstruction weight to masked patches with higher edge salience:
\begin{equation}
\mathcal{L}_{\mathrm{weight}} =
\frac{\sum_{i\in\mathcal{M}} w_i\lVert x_i-\hat{x}_i\rVert_2^2}
{\sum_{i\in\mathcal{M}}w_i},
\qquad
w_i=1+\lambda_{\mathrm{edge}}\tilde{s}_i,
\end{equation}
where $\lambda_{\mathrm{edge}}=1.0$ and $\tilde{s}_i\in[0,1]$ is the normalized patch-level edge salience.

\subsubsection{Dual RGB-Edge Reconstruction}

A shared ViT encoder is paired with separate RGB and edge decoders. Both reconstruct masked patches, and the combined objective is
\begin{equation}
\mathcal{L}_{\mathrm{dual}} =
\mathcal{L}_{\mathrm{pix}}+
\lambda_{\mathrm{edge}}\mathcal{L}_{\mathrm{sobel}}.
\end{equation}
This tests whether reconstructing wedge-boundary information improves the shared encoder.

\subsubsection{Part-Bottleneck MAE}

To test whether compact part-based representations are beneficial, we pass visible encoder tokens through a Slot-Attention bottleneck with $K=8$ slots and three iterations before decoding. The decoder accesses the slots rather than the original encoder patch tokens, forcing reconstruction information through this compact representation. Full architectural details are provided in the supplementary material.
\section{Experimental Setup}
\label{sec:experimental_setup}

\subsection{Dataset Curation and Downstream Benchmark}

Our annotated dataset consists of 9,877 high-resolution cuneiform tablet images and corresponding expert-annotated sign bounding boxes from the Electronic Babylonian Library (eBL). From these annotations, we extracted 180,759 verified sign crops spanning 15 historical periods. For our classification task, each target class is the identity of a sign (which may have multiple readings). We use this annotated collection to construct the downstream sign-recognition benchmark.

For the downstream benchmark, we retain 218 well-represented sign classes and enforce tablet-disjoint training, validation, and test partitions. The resulting splits contain 105,872 training images from 5,598 tablets, 25,988 validation images from 1,292 tablets, and 39,419 test images from 1,723 tablets.

To avoid self-supervised exposure to evaluation tablets, the CuneiMAE pretraining corpus is restricted to training-tablet data and an additional unlabeled collection that is disjoint from all downstream validation and test tablets. It contains 105,872 annotated crops from the downstream training tablets and 835,693 unlabeled sign crops extracted from a large-scale collection of tablet fragments~\cite{che2026automated}{; details of the crop extraction procedure are provided in the Supplementary Material}. {This gives an initial} total of 941,565 pretraining images. {As part of dataset curation, we compared the CSDD crops \cite{dencker2020deep} with these pretraining images and identified 963 overlapping crops. After removing them, 940,602 crops remained, which we split into pretraining training and validation sets.} No crop or parent tablet from the downstream validation or test partitions is included in this corpus. Sign-class labels are used only for downstream supervised training and evaluation.

\subsection{Evaluation Protocols}
\subsubsection{Frozen Linear Probing}
We first assess the linear separability of the pretrained representations.
The ViT encoder is frozen, and only a newly initialized classification
head is trained on the labeled dataset. Following standard linear-probing
practice, {we use the encoder's [CLS] token by default.
For the Part Bottleneck variant in Table~\ref{tab:variants}, we instead
average the encoder's patch tokens, excluding [CLS].} The resulting
features pass through a non-affine batch-normalization layer before
entering the linear classifier.

\subsubsection{Full and Label-Efficient Fine-Tuning}
We next evaluate the pretrained models under end-to-end supervised fine-tuning, where both the encoder and classification head are optimized. To evaluate label efficiency, fine-tuning is repeated using 1\%, 25\%, 50\%, 75\%, and 100\% of the available labeled training data. For each supervision level, identical subsets are evaluated across all models. This protocol evaluates how the benefit of domain-specific pretraining changes with increasing supervision.

\subsubsection{Frozen k-Nearest-Neighbor Evaluation}
To assess the organization of the learned feature space without fitting a classifier, we perform k-nearest-neighbor classification directly on frozen encoder embeddings. All embeddings are L2-normalized, and cosine similarity is used for nearest-neighbor search.

Training embeddings serve as the reference gallery, while test embeddings are used as queries. For each query, the labels of its k nearest neighbors are combined by majority voting to obtain the predicted sign class across $k \in \{1,5,10,15,20\}$.

\subsubsection{Cross-Period Transfer}
A central challenge in cuneiform recognition is the visual evolution of signs across historical periods. We therefore evaluate representation transfer across Old Babylonian (OB), Middle Babylonian (MB), and Neo-Assyrian (NA) periods. Period labels are assigned per tablet from eBL metadata.

For each source-target pair, we retain only sign identities shared by both periods. In frozen linear transfer, the encoder remains fixed while a linear classifier is trained on the source period and evaluated on the target period. We consider four directions: OB$\rightarrow$MB, MB$\rightarrow$OB, OB$\rightarrow$NA, and NA$\rightarrow$OB.

We additionally perform cross-period $k$-NN evaluation, using source-period embeddings as the reference gallery and target-period embeddings as queries, without optimizing any classifier parameters. Because the self-supervised corpus contains unlabeled images from all evaluated periods, these experiments measure cross-period transfer after domain-specific pretraining rather than generalization to a previously unseen historical period.

\begin{figure*}
    \centering
    \includegraphics[width=0.88\linewidth]{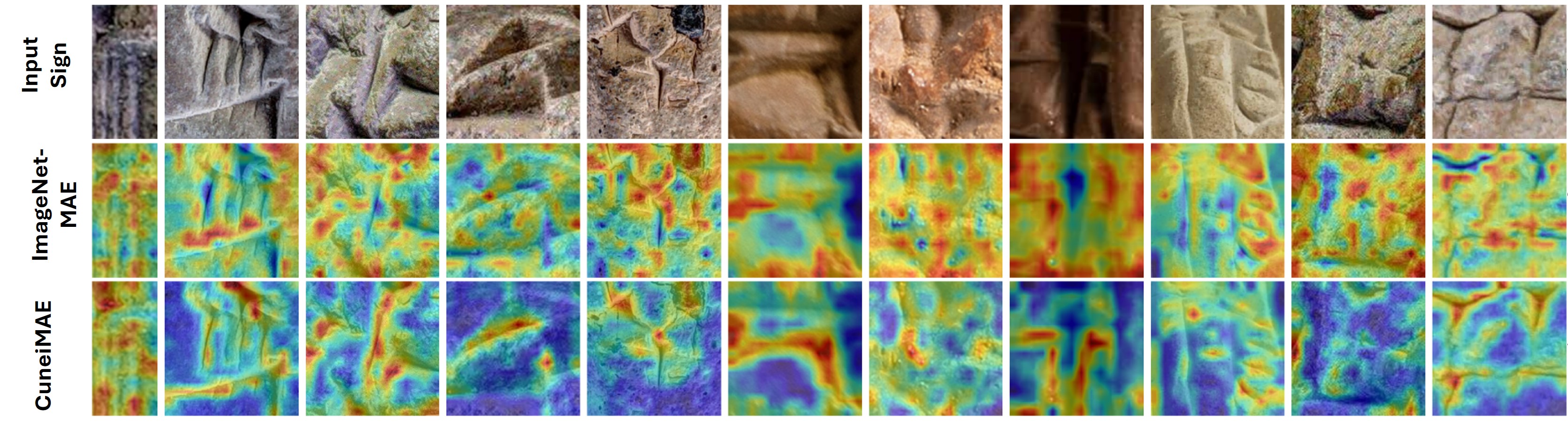}
    \caption{Qualitative comparison of spatial response maps on held-out cuneiform sign crops. From top to bottom, the rows show input crops, ImageNet-MAE heatmaps, and CuneiMAE heatmaps. Each heatmap is computed from the $\ell_2$ norms of the final transformer block's patch tokens. The resulting $14 \times 14$ grid is bilinearly upsampled to $224 \times 224$ and min--max normalized separately for each crop and encoder.}
    \label{fig:heatmap}
\end{figure*}

\subsubsection{{CSDD-Derived External Evaluation}}
\label{sec:csdd_protosnap}

{We evaluate CuneiMAE on a reconstructed CSDD-derived benchmark \cite{dencker2020deep} containing 187 sign classes, with 3,611 training, 919 validation, and 3,442 held-out test crops. Rare classes are defined as those with fewer than 50 training examples, following ProtoSnap \cite{mikulinsky2025protosnap}.} {The reconstructed split uses CSDD's official tablet-level partition with all currently available tablet images. ProtoSnap's original 180-class split is documented in its public repository; all experiments reported here use the reconstructed partitions.}

\paragraph{Synthetic-data augmentation.}
We follow the ProtoSnap-style synthetic augmentation procedure on the reconstructed CSDD-derived benchmark and evaluate its effect for
{the ResNet-18 baseline and both CuneiMAE checkpoints
(from-scratch and ImageNet-initialized)}. Each model is trained on
real CSDD-derived data alone and with additional samples generated
using ProtoSnap's diffusion-model (SD-{\cuneiformA}) and ControlNet
(CN-{\cuneiformA}) pipelines. {Both augmented settings
use the same staged training procedure.}

\paragraph{Full fine-tuning comparison.}
We additionally evaluate end-to-end sign classification using
ResNet-18~\citep{he2016deep}, ViT-B/16~\cite{dosovitskiy2021an},
ImageNet-pretrained MAE~\cite{he2022masked}, and
{both CuneiMAE checkpoints}. {All models
are fine-tuned for 90 epochs without synthetic augmentation.}
Model selection is performed on the validation split before evaluation
on the held-out test set.

{For all external evaluations, we report overall accuracy, rare-class accuracy, balanced accuracy, and rare-class balanced accuracy.}

\subsection{Implementation Details}

All domain-specific MAE variants use a ViT-B/16 backbone. {We pretrain CuneiMAE from scratch for 100 epochs in the main experiments and for 10 epochs in the ablation study, where all structure-aware variants follow the same 10-epoch protocol. These runs use AdamW} ($\mathrm{lr}=1.5\times10^{-4}$, weight decay $=0.05$). {For the external CSDD-derived benchmark only, we also evaluate a checkpoint initialized from ImageNet-MAE and further pretrained on cuneiform for 39 epochs. We selected this checkpoint because its validation reconstruction loss was close to that of the 100-epoch checkpoint pretrained from scratch.} Standard CuneiMAE uses 75\% random masking and normalized-pixel reconstruction. The structure-aware variants modify the objective, masking policy, or architecture as described in Sec.~\ref{sec:method}. For the 218-class benchmark, frozen linear probes use the \texttt{[CLS]} representation by default with a batch-normalization layer and a linear classifier, trained for 90 epochs using LARS. {All internal experiments use direct resizing to $224\times224$, without random or center cropping.} Horizontal flipping is omitted because reflection changes cuneiform sign geometry. Full optimization and preprocessing settings are provided in the supplementary material.

\section{Results and Analysis}

\subsection{Domain-Specific vs.\ Generic Visual Pretraining}
\label{sec:results_domain}

We first evaluate the linear separability of representations learned under generic visual pretraining versus domain-specific pretraining on cuneiform crops. In addition to ImageNet-MAE, we evaluate CLIP (ViT-B/16)~\citep{radford2021learning} and DINOv2 (ViT-B/14)~\citep{oquab2023dinov2} as frozen representation baselines using the same dataset splits and downstream linear-probe protocol. As reported in Table~\ref{tab:linear_probing}, CuneiMAE improves Top-1 accuracy over ImageNet-MAE by {18.45} percentage points (pp; {43.66\%} vs.\ {25.21\%}). Among the generic pretrained baselines, {ImageNet-MAE performs best in Top-1 accuracy (25.21\%), whereas DINOv2 achieves the highest Top-5 accuracy (50.93\%), outperforming ImageNet-MAE (48.45\%). CLIP underperforms ImageNet-MAE in both metrics.}

Consistent with this result, Fig.~\ref{fig:heatmap} shows that the ImageNet-MAE linear probe often produces broader responses across the sign image surface, whereas CuneiMAE tends to produce more localized responses around wedge impressions and sign boundaries. These maps are qualitative visualizations and do not constitute a localization evaluation.

\begin{table}[h]
\centering
\caption{Frozen linear probing on the 218-class test set. Best results in \textbf{bold}; second-best \underline{underlined}.}
\label{tab:linear_probing}
\begin{tabular}{lcc}
\toprule
\textbf{Pretraining Strategy} & \textbf{Top-1 (\%)} & \textbf{Top-5 (\%)} \\
\midrule
Majority-class predictor & 4.08 & -- \\
ImageNet-MAE & {\underline{25.21}} & {{48.45}} \\
CLIP (ViT-B/16) & {20.45} & {44.09} \\
DINOv2 (ViT-B/14) & {{24.93}} & \underline{{50.93}} \\
CuneiMAE & \textbf{{43.66}} & \textbf{{66.88}} \\
\bottomrule
\end{tabular}
\end{table}

\subsection{{Do Explicit Structural Priors Help?}}
\label{sec:results_priors}

We next compare the five structure-aware variants (\Cref{sec:structure}) against standard CuneiMAE.

\begin{table}[h]
\centering
{
\caption{Comparison of structure-aware pretraining variants under frozen linear probing on the test split. Best results in \textbf{bold}; second-best \underline{underlined}. {All models in this comparison were pretrained for 10 epochs.}}
\label{tab:variants}
\small
\setlength{\tabcolsep}{8pt}
\begin{tabular}{lcc}
\toprule
\textbf{Variant} & \textbf{Top-1 (\%)} & \textbf{Top-5 (\%)} \\
\midrule
Edge Loss & ${{31.32}}$ & ${55.98}$ \\
Dual Target & ${{29.72}}$ & ${53.71}$ \\
Edge40 Masking & ${21.31}$ & ${44.28}$ \\
Edge75 Masking & $\underline{{31.86}}$ & $\underline{{56.62}}$ \\
Part Bottleneck ($K=8$) & ${24.56}$ & ${48.84}$ \\
\textbf{{CuneiMAE-10}} & $\mathbf{{34.74}}$ & $\mathbf{{59.06}}$ \\
\bottomrule
\end{tabular}}
\end{table}

{As shown in Table~\ref{tab:variants}, {Edge Loss and Edge75 Masking remain relatively close to CuneiMAE, while Dual Target performs somewhat lower. Edge75 achieves the highest Top-5 accuracy among the structural variants, although it does not match CuneiMAE.} Edge40 and Part Bottleneck substantially reduce linear separability. The Edge40 variant has a much lower effective masking ratio of approximately 15.8\%, which leaves considerably more of the image visible. In contrast, Edge75 retains the standard 75\% masking ratio but still does not outperform CuneiMAE. Similarly, the Part Bottleneck variant suggests that forcing high-frequency spatial tokens through {a $K=8$ slot bottleneck} may discard fine-grained spatial information required to separate visually similar cuneiform classes.}

{These results suggest that, under our experimental setting, standard high-ratio masked pixel reconstruction is sufficient to learn useful cuneiform representations, while the evaluated hand-crafted structural constraints provide no additional benefit.
}

\subsection{Label-Efficient Fine-Tuning}
\label{sec:results_fine_tuning}

In digital epigraphy, obtaining sign annotations from Assyriologists is an expensive bottleneck. To evaluate performance under severe label scarcity, we compare ImageNet-initialized ResNet-18, ImageNet-MAE and CuneiMAE across subsets containing 1\%, 25\%, 50\%, 75\%, and 100\% of the available training labels.

\begin{table}[h]
\centering
\caption{Supervised fine-tuning Top-1 test accuracy (\%) across varying train set fractions (\%Train; out of 105,872 total images). Reported gains ($\Delta$ Gain; in pp) are with respect to ImageNet-MAE (MAE). ResNet-18 is pretrained on ImageNet. Best {model} results in \textbf{bold}; second-best {model} \underline{underlined}.}
\label{tab:label_efficiency}
\small
\setlength{\tabcolsep}{3pt}
\begin{tabular}{lcccc}
\toprule
\textbf{\%Train} & \textbf{ResNet-18} & \textbf{MAE} & \textbf{CuneiMAE} & \textbf{$\Delta$ Gain} \\
\midrule
1\%
& $15.65$ & $\underline{23.09}$ & $\mathbf{43.53}$ & +20.44 \\
25\%
& $71.94$ & $\underline{75.31}$ & $\mathbf{79.26}$ & +3.95 \\
50\% 
& $75.70$ & $\underline{80.01}$ & $\mathbf{82.70}$ & +2.69 \\
75\%
& $77.94$ & $\underline{82.27}$ & $\mathbf{84.45}$ & +2.18 \\
100\%
& $79.63$ & $\underline{83.93}$ & $\mathbf{85.49}$ & +1.56 \\
\bottomrule
\end{tabular}
\end{table}




The results in Table~\ref{tab:label_efficiency} demonstrate that domain-specific pretraining provides its highest utility in data-starved regimes. {At just 1\% supervision, CuneiMAE yields a 20.44 pp improvement over ImageNet-MAE and a 27.88 pp improvement over ImageNet-initialized ResNet-18. The gap between CuneiMAE and ImageNet-MAE narrows as supervision increases, to 1.56 pp at 100\% supervision.} CuneiMAE also consistently outperforms ResNet-18 across all label fractions.

\subsection{Neighborhood Structure of Frozen Representations}
\label{sec:results_knn}

To evaluate representation quality without optimizing any downstream classification parameters, we examine the local manifold structure using non-parametric $k$-nearest-neighbor ($k$-NN) classification on frozen, $L_2$-normalized $[\mathrm{CLS}]$ embeddings.

\begin{table}[h]
\centering
\caption{Frozen $k$-NN classification Top-1 accuracy over varying neighborhoods $k$ on the held-out test split (218 classes). Best results in \textbf{bold}.}
\label{tab:knn_eval}
\small
\setlength{\tabcolsep}{5pt}
\resizebox{\linewidth}{!}{
\begin{tabular}{lccccc}
\toprule
\textbf{Backbone} & \textbf{$k=1$} & \textbf{$k=5$} & \textbf{$k=10$} & \textbf{$k=15$} & \textbf{$k=20$} \\
\midrule
{ImageNet-MAE} & {6.07} & {6.91} & {7.79} & {8.03} & {8.35} \\
{CuneiMAE} & \textbf{{13.74}} & \textbf{{14.83}} & \textbf{{16.29}} & \textbf{{16.88}} & \textbf{{16.99}} \\
\midrule
{$\Delta$ (pp)} & {+7.67} & {+7.92} & {+8.50} & {\textbf{+8.85}} & {+8.64} \\
\bottomrule
\end{tabular}
}
\end{table}

Across all choices of $k$, CuneiMAE embeddings yield substantially higher $k$-NN classification accuracy. As shown in Table~\ref{tab:knn_eval}, {CuneiMAE achieves 16.29\% Top-1 accuracy at $k=10$, more than doubling the 7.79\% achieved by ImageNet-MAE, for a gain of 8.50 percentage points.} CuneiMAE’s advantage holds across all neighborhood sizes.
To complement the quantitative $k$-NN results, qualitative top-3 nearest-neighbor retrieval examples are provided in the supplementary material. These examples include both successes and failures, providing additional insight into the local structure of the frozen embedding spaces.

\subsection{Cross-Period Transfer}
\label{sec:results_cross_period}

We evaluate cross-period transfer across OB, MB, and NA using the protocol described in Sec.~\ref{sec:experimental_setup}.

As shown in Table~\ref{tab:cross_period_linear}, {domain-specific pretraining consistently outperforms ImageNet-MAE across all evaluated temporal directions by 3.25--10.80 pp.}

\begin{table}[h]
\centering
\caption{Cross-period transfer Top-1 accuracy (\%) under frozen linear probing. Training images are drawn from the source period (left-hand side of ``Transfer Split'') and images being evaluated are drawn from the target period (right-hand side). MAE refers to ImageNet-MAE. Best results in \textbf{bold}.}
\label{tab:cross_period_linear}
\small
\setlength{\tabcolsep}{4pt}
\begin{tabular}{lccc}
\toprule
\textbf{Transfer Split} & \textbf{MAE} & \textbf{CuneiMAE} & \textbf{$\Delta$ (pp)} \\
\midrule
OB $\rightarrow$ MB & {10.25} & \textbf{{21.05}} & {+10.80} \\
MB $\rightarrow$ OB & {10.71} & \textbf{{17.79}} & {+7.08} \\
OB $\rightarrow$ NA & {9.10} & \textbf{{14.57}} & {+5.47} \\
NA $\rightarrow$ OB & {9.72} & \textbf{{12.97}} & {+3.25} \\
\bottomrule
\end{tabular}
\end{table}

Cross-period $k$-NN evaluation provides complementary evidence from the frozen embedding space.

\begin{table}[h]
\centering
{
\caption{Cross-period non-parametric $k$-NN evaluation ($k=10$) Top-1 accuracy (\%). Potential neighbor images are drawn from the source period (left-hand side of ``Transfer Split'') and query images being evaluated are drawn from the target period (right-hand side). Majority predicts the most frequent source-period class for all queries. MAE refers to ImageNet-MAE; $\Delta$ is relative to MAE. Best results in \textbf{bold}.}
\label{tab:cross_period_knn}
\small
\setlength{\tabcolsep}{4pt}
\begin{tabular}{lcccc}
\toprule
\textbf{Transfer Split} & \textbf{Majority} & \textbf{MAE} & \textbf{CuneiMAE} & \textbf{$\Delta$ (pp)} \\
\midrule
OB $\rightarrow$ MB & {3.24} & {3.70} & \textbf{{6.86}} & {+3.16} \\
MB $\rightarrow$ OB & 2.21 & {3.40} & \textbf{{7.52}} & {+4.12} \\
OB $\rightarrow$ NA & 4.70 & {3.30} & \textbf{{5.04}} & {+1.74} \\
NA $\rightarrow$ OB & 5.76 & {3.97} & \textbf{{6.17}} & {+2.20} \\
\bottomrule
\end{tabular}
}
\end{table}

Table~\ref{tab:cross_period_knn} shows that ImageNet-MAE features score above the majority-class baseline in two transfer directions but fall below it in the other two, indicating limited sign-discriminative neighborhood structure across periods. {CuneiMAE exceeds this baseline in all four directions, by 3.62 and 5.31 pp for OB/MB transfers and by 0.34 and 0.41 pp for transfers involving NA.} Absolute accuracy remains low, consistent with the substantial differences in morphology between OB and NA sign forms which often bear little visual similarity.

\subsection{{CSDD-Derived External Evaluation}}
\label{sec:results_csdd}

We evaluate ProtoSnap-style staged training and synthetic augmentation on the reconstructed 187-class CSDD-derived benchmark described in Sec.~\ref{sec:csdd_protosnap}, using direct resizing to $224\times224$ without cropping.  A supplementary preprocessing ablation examines the effects of crop-and-flip preprocessing {relative to direct resizing and reports additional direct-resizing results for CuneiMAE checkpoints pretrained for 10 and 27 epochs}.

\begin{table}[h]
\centering
\caption{Sign-classification performance on the reconstructed 187-class
CSDD-derived test split (\%)~\cite{dencker2020deep} without cropping.
CSDD denotes training on real data only.
SD-{\cuneiformA} and CN-{\cuneiformA} use synthetic samples from the released
ProtoSnap diffusion and ControlNet models, respectively.
Augmented runs use 8 epochs of training on real and synthetic data,
followed by 10 epochs of fine-tuning on real data.
Bal.\ denotes balanced accuracy.
Bold marks the best result for each metric within each model block.}
\label{tab:external_sign_classification}
\small
\setlength{\tabcolsep}{4pt}
\begin{tabular}{lcccc}
\toprule
\textbf{Model / Setting} & \textbf{Acc.} & \textbf{Rare Acc.} &
\textbf{Bal.} & \textbf{Rare Bal.} \\
\midrule
\multicolumn{5}{l}{\textit{ResNet-18}} \\
\hspace{0.5em} CSDD & 58.72 & 46.14 & 37.08 & 32.90 \\
\hspace{0.5em} +SD-{\cuneiformA} & 62.99 & \textbf{52.66} &
\textbf{40.86} & \textbf{37.04} \\
\hspace{0.5em} +CN-{\cuneiformA} & \textbf{63.80} & 52.41 & 40.54 & 36.48 \\
\midrule
\multicolumn{5}{l}{{\textit{CuneiMAE (scratch)}}} \\
\hspace{0.5em} CSDD & {67.49} & {57.29} &
{43.32} & {39.27} \\
\hspace{0.5em} +SD-{\cuneiformA} & {\textbf{69.05}} &
{\textbf{60.47}} & {\textbf{48.54}} &
{\textbf{45.22}} \\
\hspace{0.5em} +CN-{\cuneiformA} & {67.78} &
{58.64} & {47.67} & {44.24} \\
\midrule
\multicolumn{5}{l}{{\textit{CuneiMAE (ImageNet init.)}}} \\
\hspace{0.5em} CSDD & {68.74} & {60.63} &
{50.21} & {47.06} \\
\hspace{0.5em} +SD-{\cuneiformA} & {\textbf{70.48}} &
{\textbf{62.37}} & {\textbf{50.55}} &
{\textbf{47.28}} \\
\hspace{0.5em} +CN-{\cuneiformA} & {67.93} &
{59.33} & {48.82} & {45.66} \\
\bottomrule
\end{tabular}
\end{table}

\begin{table}[t]
\centering
\caption{Sign-classification performance on the reconstructed 187-class
CSDD-derived test split (\%) after 90 epochs of end-to-end fine-tuning
without synthetic augmentation. Checkpoints are selected on the
validation split. Bal.\ denotes balanced accuracy.
For each metric, the best result is in \textbf{bold} and the second-best
is \underline{underlined}.}
\label{tab:external_sign_classification_finetuned}
\small
\setlength{\tabcolsep}{4pt}
\resizebox{\linewidth}{!}{
\begin{tabular}{lcccc}
\toprule
\textbf{Model} & \textbf{Acc.} & \textbf{Rare Acc.} &
\textbf{Bal.} & \textbf{Rare Bal.} \\
\midrule
ResNet-18 & 61.97 & 49.93 & 39.30 & 35.27 \\
ViT-B/16~\cite{dosovitskiy2021an} & 58.02 & 47.39 & 35.16 & 31.39 \\
ImageNet-MAE~\cite{he2022masked} & 56.54 & 44.65 & 33.35 & 29.19 \\
{CuneiMAE (scratch)} &
{\underline{67.55}} &
{\underline{58.44}} &
{\underline{47.90}} &
{\underline{44.54}} \\
{CuneiMAE (ImageNet init.)} &
{\textbf{69.87}} &
{\textbf{60.78}} &
{\textbf{51.82}} &
{\textbf{48.60}} \\
\bottomrule
\end{tabular}
}
\end{table}

Table~\ref{tab:external_sign_classification} reports sign-classification
performance for the ImageNet-pretrained ResNet-18 baseline and
{both CuneiMAE checkpoints}. The real-only ResNet-18 baseline achieves 58.72\% accuracy on the reconstructed 187-class split. {For reference, ProtoSnap~\cite{mikulinsky2025protosnap} reported 58.43\% on its original 180-class split.} {SD-{\cuneiformA} improves all four metrics for ResNet-18 and both CuneiMAE checkpoints. CN-{\cuneiformA} improves all four metrics for ResNet-18 and the from-scratch CuneiMAE checkpoint but lowers them for the ImageNet-initialized checkpoint relative to its real-only baseline. Among the ResNet-18 settings, CN-{\cuneiformA} augmentation achieves the highest accuracy (63.80\%). Across all evaluated settings, ImageNet-initialized CuneiMAE with SD-{\cuneiformA} augmentation achieves the best results on all four metrics: 70.48\% accuracy, 62.37\% rare
accuracy, 50.55\% balanced accuracy, and 47.28\% rare balanced accuracy.}

{Table~\ref{tab:external_sign_classification_finetuned}
further compares end-to-end fine-tuning across ImageNet-pretrained
ResNet-18 and ViT-B/16, ImageNet-MAE, and both CuneiMAE checkpoints on
the same benchmark without cropping. ImageNet-initialized CuneiMAE
achieves the highest performance across all four metrics, with 69.87\%
accuracy, compared with 67.55\% for CuneiMAE pretrained from scratch and
61.97\% for ResNet-18. Training details are provided in the supplementary
material.}

\section{Discussion}

Our results reveal a clear distinction between frozen representation quality and downstream adaptation. CuneiMAE provides stronger frozen features, while its advantage decreases as the amount of labeled data used for end-to-end fine-tuning increases. This suggests that domain-specific self-supervision is particularly valuable when downstream annotations are limited. {CuneiMAE also outperforms ResNet-18 at every label fraction on the internal benchmark (Table~\ref{tab:label_efficiency}). On the external CSDD-derived benchmark, both CuneiMAE checkpoints outperform ImageNet-pretrained ViT-B/16 and MAE, as well as ResNet-18, across all four metrics after fine-tuning without synthetic data. The ImageNet-initialized checkpoint performs best, but the two checkpoints differ in pretraining duration, so this comparison does not isolate the effect of initialization. Both runs averaged approximately 1 hour and 40 minutes per epoch, corresponding to about 65 hours of in-domain pretraining for the 39-epoch ImageNet-initialized checkpoint and 167 hours for the 100-epoch checkpoint trained from scratch. The pretraining loss curves are shown in the supplementary material.} The benchmarks differ in source collection {and} image acquisition{, while our CSDD ablation shows that preprocessing choices can affect models unequally}. These differences motivate further investigation of how preprocessing and dataset characteristics influence the benefits of domain-specific pretraining. Future work could also explore CuneiMAE-style self-supervised pretraining with convolutional backbones to assess whether domain-specific representation learning can benefit CNN architectures.

ProtoSnap and CuneiMAE address complementary aspects of cuneiform analysis. ProtoSnap recovers wedge-level structure through prototype alignment and uses it to generate synthetic training data, whereas CuneiMAE learns general-purpose representations through masked-image reconstruction. Our synthetic-data experiments indicate that the two can be combined, {but the effect of augmentation depends on both the generator and the CuneiMAE checkpoint.}

The structure-aware experiments indicate that the evaluated hand-crafted priors provide no clear advantage over standard CuneiMAE {under the controlled 10-epoch pretraining setting}. The contrast between Edge40 and Edge75 highlights the importance of controlling the masking ratio when evaluating structure-guided masking. A plausible reason for the limited benefit of edge priors is that Sobel gradients in photographs of clay reflect lighting-dependent shading as well as wedge geometry, and breaks, cracks, ruling lines, and tablet edges also produce strong responses. Future work could test whether structural priors are more effective with 3D scans or multi-light captures.


\section{Conclusion}

We investigated domain-specific self-supervised learning for cuneiform sign representation using masked autoencoding. In-domain pretraining consistently outperformed generic pretrained baselines in frozen, low-label, and cross-period evaluations on our internal benchmark, with the largest gains when labeled data were limited, highlighting the value of domain-specific pretraining for low-resource historical document analysis. On the smaller external CSDD-derived benchmark, {both CuneiMAE checkpoints outperformed ResNet-18 under full fine-tuning without synthetic data, with the ImageNet-initialized checkpoint achieving the highest accuracy. In a separate staged-training protocol, ProtoSnap's diffusion-generated data improved accuracy for both CuneiMAE checkpoints.} Our structure-aware variants did not improve over standard pixel reconstruction {under controlled 10-epoch pretraining}, suggesting that the edge- and part-based constraints examined here are not necessary in this setting. 
{Overall, domain-specific masked image modeling is an effective approach for cuneiform representation learning, especially when labels are scarce.}
\section*{Acknowledgment}
This research is part of the project RECC (Rewriting the End of Cuneiform Culture), funded by the European Research Council (ERC), Grant Agreement No.~101171038 (\url{https://doi.org/10.3030/101171038})
{
    \small
    \bibliographystyle{ieeenat_fullname}
    \bibliography{main}

@String(CVPR= {IEEE Conf. Comput. Vis. Pattern Recog.})

@String(ICCV= {Int. Conf. Comput. Vis.})

@String(CVPR  = {CVPR})

@String(ICCV  = {ICCV})

@inproceedings{kriege2018recognizing,
  title={Recognizing cuneiform signs using graph based methods},
  author={Kriege, Nils M and Fey, Matthias and Fisseler, Denis and Mutzel, Petra and Weichert, Frank},
  booktitle={International workshop on cost-sensitive learning},
  pages={31--44},
  year={2018},
  organization={PMLR}
}

@article{dencker2020deep,
  title={Deep learning of cuneiform sign detection with weak supervision using transliteration alignment},
  author={Dencker, Tobias and Klinkisch, Pablo and Maul, Stefan M and Ommer, Bj{\"o}rn},
  journal={Plos one},
  volume={15},
  number={12},
  pages={e0243039},
  year={2020},
  publisher={Public Library of Science San Francisco, CA USA}
}

@inproceedings{
dosovitskiy2021an,
title={An Image is Worth 16x16 Words: Transformers for Image Recognition at Scale},
author={Alexey Dosovitskiy and Lucas Beyer and Alexander Kolesnikov and Dirk Weissenborn and Xiaohua Zhai and Thomas Unterthiner and Mostafa Dehghani and Matthias Minderer and Georg Heigold and Sylvain Gelly and Jakob Uszkoreit and Neil Houlsby},
booktitle={International Conference on Learning Representations},
year={2021},
url={https://openreview.net/forum?id=YicbFdNTTy}
}

@article{williams2025deepscribe,
  title={DeepScribe: localization and classification of elamite cuneiform signs via deep learning},
  author={Williams, Edward C and Su, Grace and Schloen, Sandra R and Prosser, Miller and Paulus, Susanne and Krishnan, Sanjay},
  journal={ACM Journal on Computing and Cultural Heritage},
  volume={18},
  number={2},
  pages={1--32},
  year={2025},
  publisher={ACM New York, NY}
}

@article{cobanoglu2024sign,
  title={Sign detection for cuneiform tablets},
  author={Cobanoglu, Yunus and S{\'a}enz, Luis and Khait, Ilya and Jim{\'e}nez, Enrique},
  journal={it-Information Technology},
  volume={66},
  number={1},
  pages={28--38},
  year={2024},
  publisher={De Gruyter Oldenbourg}
}

@inproceedings{he2022masked,
  title={Masked autoencoders are scalable vision learners},
  author={He, Kaiming and Chen, Xinlei and Xie, Saining and Li, Yanghao and Doll{\'a}r, Piotr and Girshick, Ross},
  booktitle={2022 IEEE/CVF conference on computer vision and pattern recognition (CVPR)},
  pages={15979--15988},
  year={2022},
  organization={IEEE}
}

@article{bogacz2022digital,
  title={Digital assyriology—advances in visual cuneiform analysis},
  author={Bogacz, Bartosz and Mara, Hubert},
  journal={Journal on Computing and Cultural Heritage (JOCCH)},
  volume={15},
  number={2},
  pages={1--22},
  year={2022},
  publisher={ACM New York, NY}
}

@article{lastilla2022self,
  title={Self-supervised learning for medieval handwriting identification: A case study from the Vatican Apostolic Library},
  author={Lastilla, Lorenzo and Ammirati, Serena and Firmani, Donatella and Komodakis, Nikos and Merialdo, Paolo and Scardapane, Simone},
  journal={Information Processing \& Management},
  volume={59},
  number={3},
  pages={102875},
  year={2022},
  publisher={Elsevier}
}

@article{jemni2026st,
  title={ST-KeyS: Self-supervised Transformer for Keyword Spotting in historical handwritten documents},
  author={Jemni, Sana Khamekhem and Ammar, Sourour and Souibgui, Mohamed Ali and Kessentini, Yousri and Cheddad, Abbas},
  journal={Pattern Recognition},
  volume={170},
  pages={112036},
  year={2026},
  publisher={Elsevier}
}

@inproceedings{mikulinsky2025protosnap,
  title={Protosnap: Prototype alignment for cuneiform signs},
  author={Mikulinsky, Rachel and Alper, Morris and Gordin, Shai and Jim{\'e}nez, Enrique and Cohen, Yoram and Averbuch-Elor, Hadar},
  booktitle={International Conference on Learning Representations},
  volume={2025},
  pages={88720--88739},
  year={2025}
}

@inproceedings{chen2024classification,
  title={Classification of paleographic artifacts at scale: Mitigating confounds and distribution shift in cuneiform tablet dating},
  author={Chen, Danlu and Tian, Jiahe and Weng, Yufei and Berg-Kirkpatrick, Taylor and Myerston, Jacobo},
  booktitle={Proceedings of the 1st Workshop on Machine Learning for Ancient Languages (ML4AL 2024)},
  pages={30--41},
  year={2024}
}

@article{rest2022illumination,
  title={Illumination-based augmentation for cuneiform deep neural sign classification},
  author={Rest, Christopher and Fisseler, Denis and Weichert, Frank and Somel, Turna and M{\"u}ller, Gerfrid GW},
  journal={Journal on Computing and Cultural Heritage (JOCCH)},
  volume={15},
  number={3},
  pages={1--20},
  year={2022},
  publisher={ACM New York, NY}
}

@article{che2026automated,
  title={Automated sign detection across the Electronic Babylonian Library: A large-scale dataset and end-to-end cuneiform OCR pipeline},
  author={Che, Wentao and Arias, Esteban Garc{\'e}s and Niaz, Asim and Bender, Andreas and Jim{\'e}nez, Enrique},
  journal={arXiv preprint arXiv:2606.22608},
  year={2026}
}

@article{kapon2024shaping,
  title={Shaping history: Advanced machine learning techniques for the analysis and dating of cuneiform tablets over three millennia},
  author={Kapon, Danielle and Fire, Michael and Gordin, Shai},
  journal={arXiv preprint arXiv:2406.04039},
  year={2024}
}

@inproceedings{li2022dit,
  title={Dit: Self-supervised pre-training for document image transformer},
  author={Li, Junlong and Xu, Yiheng and Lv, Tengchao and Cui, Lei and Zhang, Cha and Wei, Furu},
  booktitle={Proceedings of the 30th ACM international conference on multimedia},
  pages={3530--3539},
  year={2022}
}

@article{alfaro2023few,
  title={Few-shot symbol classification via self-supervised learning and nearest neighbor},
  author={Alfaro-Contreras, Mar{\'\i}a and R{\'\i}os-Vila, Antonio and Valero-Mas, Jose J and Calvo-Zaragoza, Jorge},
  journal={Pattern Recognition Letters},
  volume={167},
  pages={1--8},
  year={2023},
  publisher={Elsevier}
}

@inproceedings{stotzner2023cnn,
  title={Cnn based cuneiform sign detection learned from annotated 3d renderings and mapped photographs with illumination augmentation},
  author={St{\"o}tzner, Ernst and Homburg, Timo and Mara, Hubert},
  booktitle={2023 IEEE/CVF International Conference on Computer Vision Workshops (ICCVW)},
  pages={1672--1680},
  year={2023},
  organization={IEEE}
}

@unpublished{niaz2026tabletaware,
  author = {Niaz, Asim and Bender, Andreas and Umraiz, Muhammad and Jim{\'e}nez, Enrique},
  title  = {A Tablet-Aware Visual Analysis Framework for Cuneiform Sign Recognition, Period Attribution, and Fragment Retrieval},
  year   = {2026},
  note   = {Manuscript submitted to the Humanities and Social Sciences Communications}
}

@article{yugay2024stylistic,
  title={Stylistic classification of cuneiform signs using convolutional neural networks},
  author={Yugay, Vasiliy and Paliwal, Kartik and Cobanoglu, Yunus and S{\'a}enz, Luis and Gogokhia, Ekaterine and Gordin, Shai and Jim{\'e}nez, Enrique},
  journal={IT-Information Technology},
  volume={66},
  number={1},
  pages={15--27},
  year={2024},
  publisher={De Gruyter Oldenbourg}
}

@inproceedings{chen2020simple,
  title={A simple framework for contrastive learning of visual representations},
  author={Chen, Ting and Kornblith, Simon and Norouzi, Mohammad and Hinton, Geoffrey},
  booktitle={International conference on machine learning},
  pages={1597--1607},
  year={2020},
  organization={PmLR}
}

@inproceedings{caron2021emerging,
  title={Emerging properties in self-supervised vision transformers},
  author={Caron, Mathilde and Touvron, Hugo and Misra, Ishan and J{\'e}gou, Herv{\'e} and Mairal, Julien and Bojanowski, Piotr and Joulin, Armand},
  booktitle={2021 IEEE/CVF international conference on computer vision (ICCV)},
  pages={9630--9640},
  year={2021},
  organization={IEEE}
}

@article{cong2022satmae,
  title={Satmae: Pre-training transformers for temporal and multi-spectral satellite imagery},
  author={Cong, Yezhen and Khanna, Samar and Meng, Chenlin and Liu, Patrick and Rozi, Erik and He, Yutong and Burke, Marshall and Lobell, David and Ermon, Stefano},
  journal={Advances in Neural Information Processing Systems},
  volume={35},
  pages={197--211},
  year={2022}
}

@article{verwimp2025signs,
  title={Signs of the Past, Patterns of the Present: On the Automatic Classification of Old Babylonian Cuneiform Signs},
  author={Verwimp, Eli and Smidt, Gustav Ryberg and Hameeuw, Hendrik and De Graef, Katrien},
  journal={arXiv preprint arXiv:2507.13959},
  year={2025}
}

@article{hamplova2024cuneiform,
  title={Cuneiform Stroke Recognition and Vectorization in 2D Images.},
  author={Hamplov{\'a}, Ad{\'e}la and Romach, Avital and Pavl{\'\i}{\v{c}}ek, Josef and Vesel{\`y}, Arno{\v{s}}t and {\v{C}}ejka, Martin and Franc, David and Gordin, Shai},
  journal={DHQ: Digital Humanities Quarterly},
  volume={18},
  number={1},
  year={2024}
}

@inproceedings{stotzner2023r,
  title={R-CNN based Polygonal Wedge Detection Learned from Annotated 3D Renderings and Mapped Photographs of Open Data Cuneiform Tablets.},
  author={St{\"o}tzner, Ernst and Homburg, Timo and Bullenkamp, Jan Philipp and Mara, Hubert},
  booktitle={GCH},
  pages={47--56},
  year={2023}
}

@inproceedings{rusakov2020towards,
  title={Towards query-by-eXpression retrieval of cuneiform signs},
  author={Rusakov, Eugen and Somel, Turna and Fink, Gernot A and M{\"u}ller, Gerfrid GW},
  booktitle={2020 17th International Conference on Frontiers in Handwriting Recognition (ICFHR)},
  pages={43--48},
  year={2020},
  organization={IEEE}
}

@inproceedings{assael2019restoring,
  title={Restoring ancient text using deep learning: a case study on Greek epigraphy},
  author={Assael, Yannis and Sommerschield, Thea and Prag, Jonathan},
  booktitle={Proceedings of the 2019 conference on empirical methods in natural language processing and the 9th international joint conference on natural language processing (EMNLP-IJCNLP)},
  pages={6368--6375},
  year={2019}
}

@inproceedings{huang2019obc306,
  title={Obc306: A large-scale oracle bone character recognition dataset},
  author={Huang, Shuangping and Wang, Haobin and Liu, Yongge and Shi, Xiaosong and Jin, Lianwen},
  booktitle={2019 International Conference on Document Analysis and Recognition (ICDAR)},
  pages={681--688},
  year={2019},
  organization={IEEE}
}

@inproceedings{radford2021learning,
  title={Learning transferable visual models from natural language supervision},
  author={Radford, Alec and Kim, Jong Wook and Hallacy, Chris and Ramesh, Aditya and Goh, Gabriel and Agarwal, Sandhini and Sastry, Girish and Askell, Amanda and Mishkin, Pamela and Clark, Jack and others},
  booktitle={International conference on machine learning},
  pages={8748--8763},
  year={2021},
  organization={PmLR}
}

@article{oquab2023dinov2,
  title={Dinov2: Learning robust visual features without supervision},
  author={Oquab, Maxime and Darcet, Timoth{\'e}e and Moutakanni, Th{\'e}o and Vo, Huy and Szafraniec, Marc and Khalidov, Vasil and Fernandez, Pierre and Haziza, Daniel and Massa, Francisco and El-Nouby, Alaaeldin and others},
  journal={arXiv preprint arXiv:2304.07193},
  year={2023}
}

@inproceedings{he2016deep,
  title={Deep residual learning for image recognition},
  author={He, Kaiming and Zhang, Xiangyu and Ren, Shaoqing and Sun, Jian},
  booktitle={Proceedings of the IEEE conference on computer vision and pattern recognition},
  pages={770--778},
  year={2016}
}

@inproceedings{zhou2023self,
  title={Self pre-training with masked autoencoders for medical image classification and segmentation},
  author={Zhou, Lei and Liu, Huidong and Bae, Joseph and He, Junjun and Samaras, Dimitris and Prasanna, Prateek},
  booktitle={2023 IEEE 20th international symposium on biomedical imaging (ISBI)},
  pages={1--6},
  year={2023},
  organization={IEEE}
}

@inproceedings{yang2022reading,
  title={Reading and writing: Discriminative and generative modeling for self-supervised text recognition},
  author={Yang, Mingkun and Liao, Minghui and Lu, Pu and Wang, Jing and Zhu, Shenggao and Luo, Hualin and Tian, Qi and Bai, Xiang},
  booktitle={Proceedings of the 30th ACM International Conference on Multimedia},
  pages={4214--4223},
  year={2022}
}

@article{lyu2022maskocr,
  title={Maskocr: Text recognition with masked encoder-decoder pretraining},
  author={Lyu, Pengyuan and Zhang, Chengquan and Liu, Shanshan and Qiao, Meina and Xu, Yangliu and Wu, Liang and Yao, Kun and Han, Junyu and Ding, Errui and Wang, Jingdong},
  journal={arXiv preprint arXiv:2206.00311},
  year={2022}
}

@inproceedings{kivsvs2024self,
  title={Self-supervised pre-training of text recognizers},
  author={Ki{\v{s}}{\v{s}}, Martin and Hradi{\v{s}}, Michal},
  booktitle={International Conference on Document Analysis and Recognition},
  pages={218--235},
  year={2024},
  organization={Springer}
}

@inproceedings{vogler2022lacuna,
  title={Lacuna reconstruction: Self-supervised pre-training for low-resource historical document transcription},
  author={Vogler, Nikolai and Allen, Jonathan and Miller, Matthew and Berg-Kirkpatrick, Taylor},
  booktitle={Findings of the Association for Computational Linguistics: NAACL 2022},
  pages={206--216},
  year={2022}
}

@article{sommerschield2023machine,
  title={Machine learning for ancient languages: A survey},
  author={Sommerschield, Thea and Assael, Yannis and Pavlopoulos, John and Stefanak, Vanessa and Senior, Andrew and Dyer, Chris and Bodel, John and Prag, Jonathan and Androutsopoulos, Ion and De Freitas, Nando},
  journal={Computational Linguistics},
  volume={49},
  number={3},
  pages={703--747},
  year={2023}
}

@article{diao2025ancient,
  title={Ancient script image recognition and processing: A review},
  author={Diao, Xiaolei and Bo, Rite and Xiao, Yanling and Shi, Lida and Zhou, Zhihan and Xu, Hao and Li, Chuntao and Tang, Xiongfeng and Poesio, Massimo and John, C{\'e}dric M and others},
  journal={arXiv preprint arXiv:2506.19208},
  year={2025}
}

@inproceedings{brandenbusch2021context,
  title={Context aware generation of cuneiform signs},
  author={Brandenbusch, Kai and Rusakov, Eugen and Fink, Gernot A},
  booktitle={International Conference on Document Analysis and Recognition},
  pages={65--79},
  year={2021},
  organization={Springer}
}

@article{de2025domain,
  title={In-domain versus out-of-domain transfer learning for document layout analysis},
  author={De Nardin, Axel and Zottin, Silvia and Piciarelli, Claudio and Foresti, Gian Luca and Colombi, Emanuela},
  journal={International Journal on Document Analysis and Recognition (IJDAR)},
  volume={28},
  number={2},
  pages={161--175},
  year={2025},
  publisher={Springer}
}

@inproceedings{de2023imagenet,
  title={Is imagenet always the best option? an overview on transfer learning strategies for document layout analysis},
  author={De Nardin, Axel and Zottin, Silvia and Colombi, Emanuela and Piciarelli, Claudio and Foresti, Gian Luca},
  booktitle={International Conference on Image Analysis and Processing},
  pages={489--499},
  year={2023},
  organization={Springer}
}
}
\clearpage
\section*{Supplementary Material}

\section{{Extraction of Unlabeled Sign Crops for CuneiMAE Pretraining}}

{We generated the unlabeled pretraining crops from the large-scale eBL fragment collection described in the main text. We processed photographs of 86,040 cune{i}form tablet fragments that had never received manual sign-level annotations using DETR \cite{che2026automated}. DETR generated 1,279,146 candidate sign crops, with at least one detection in 77,442 fragments.}

{To select crops for pretraining, we applied the Swin Base sign classifier \cite{niaz2026tabletaware} to the DETR candidates. We retained a crop when the DETR detection confidence was at least 0.60, the classifier confidence was at least 0.50, and the detector and classifier predictions agreed on the sign class. This filtering yielded 835,693 crops from 73,687 fragments. The predicted sign classes were used only for crop filtering; they were not used as targets during self-supervised pretraining. The retained crops were treated as unlabeled images for CuneiMAE pretraining.}

\section{Implementation Details}
\label{supp:implementation}

This section provides the training, optimization, and preprocessing settings for the internal 218-class benchmark. External CSDD-derived evaluation protocols are described separately in Sec.~\ref{supp:protosnap_details}.

\begin{table}[h]
\centering
\caption{Additional implementation and preprocessing details for the internal 218-class benchmark. LR=learning rate; WD=weight decay.}
\label{tab:supp_implementation}
\small
\setlength{\tabcolsep}{5pt}
\renewcommand{\arraystretch}{1.15}
\begin{tabular}{lp{5.2cm}}
\toprule
\textbf{Setting} & \textbf{Configuration} \\
\midrule
\multicolumn{2}{l}{\textit{Pretraining}} \\
\hspace{0.5em} Batch / effective batch & {64 / 256 (accumulation 4)} \\
\hspace{0.5em} Mask ratio & {0.75} \\
\hspace{0.5em} LR / WD & {$1.5\times10^{-4}$ / 0.05} \\
\hspace{0.5em} Epochs / warm-up & {100 / 5} \\
\addlinespace
\multicolumn{2}{l}{\textit{Linear Probing}} \\
\hspace{0.5em} LR / WD & 0.1 / 0.0 \\
\hspace{0.5em} Batch size & {64} \\
\hspace{0.5em} Train transform & {Resize to $224\times224$ (bicubic)} \\
\hspace{0.5em} Val. / test transform & {Same as train} \\
\addlinespace
\multicolumn{2}{l}{\textit{Normalization}} \\
\hspace{0.5em} Method & ImageNet mean/std \\
\bottomrule
\end{tabular}
\end{table}

Edge40 and Edge75 replace the default random masking policy as described in the Edge-Constrained Masking subsection of the main paper. All remaining variants use the default 75\% masking ratio unless stated otherwise.

\subsection{Label-Efficient Fine-Tuning}
\label{supp:label_efficiency}

Table~\ref{tab:supp_label_efficiency} summarizes the
fine-tuning configurations for the internal 218-class
benchmark.

\begin{table}[h]
\centering
\caption{Label-efficiency fine-tuning settings on the internal 218-class benchmark. Both MAE models share identical settings and differ only in initialization.}
\label{tab:supp_label_efficiency}
\small
\setlength{\tabcolsep}{3pt}
\renewcommand{\arraystretch}{1.15}
\begin{tabular}{l p{2.7cm} c}
\toprule
\textbf{Setting}
& \textbf{MAE (ViT-B/16)}
& \textbf{ResNet-18} \\
\midrule
Initialization & CuneiMAE, ImageNet-MAE & {ImageNet-1K} \\
Optimizer      & Adam & {Adam} \\
Epochs         & 100 & {100} \\
Batch size     & {32 (effective 64)} & {64} \\
Accumulation steps & {2} & {--} \\
Base LR / LR   & $1\times10^{-4}$ & {$5\times10^{-4}$} \\
Warm-up epochs & 5 & {0} \\
\bottomrule
\end{tabular}
\end{table}


\noindent\textbf{Preprocessing.}
{All models use direct resizing to $224\times224$, followed by ImageNet normalization. No random cropping, center cropping, or horizontal flipping is applied.}

\subsection{CuneiMAE Pretraining Convergence}
\label{supp:cuneimae_pretraining}

\begin{figure*}[t]
    \centering
    \includegraphics[width=\linewidth]{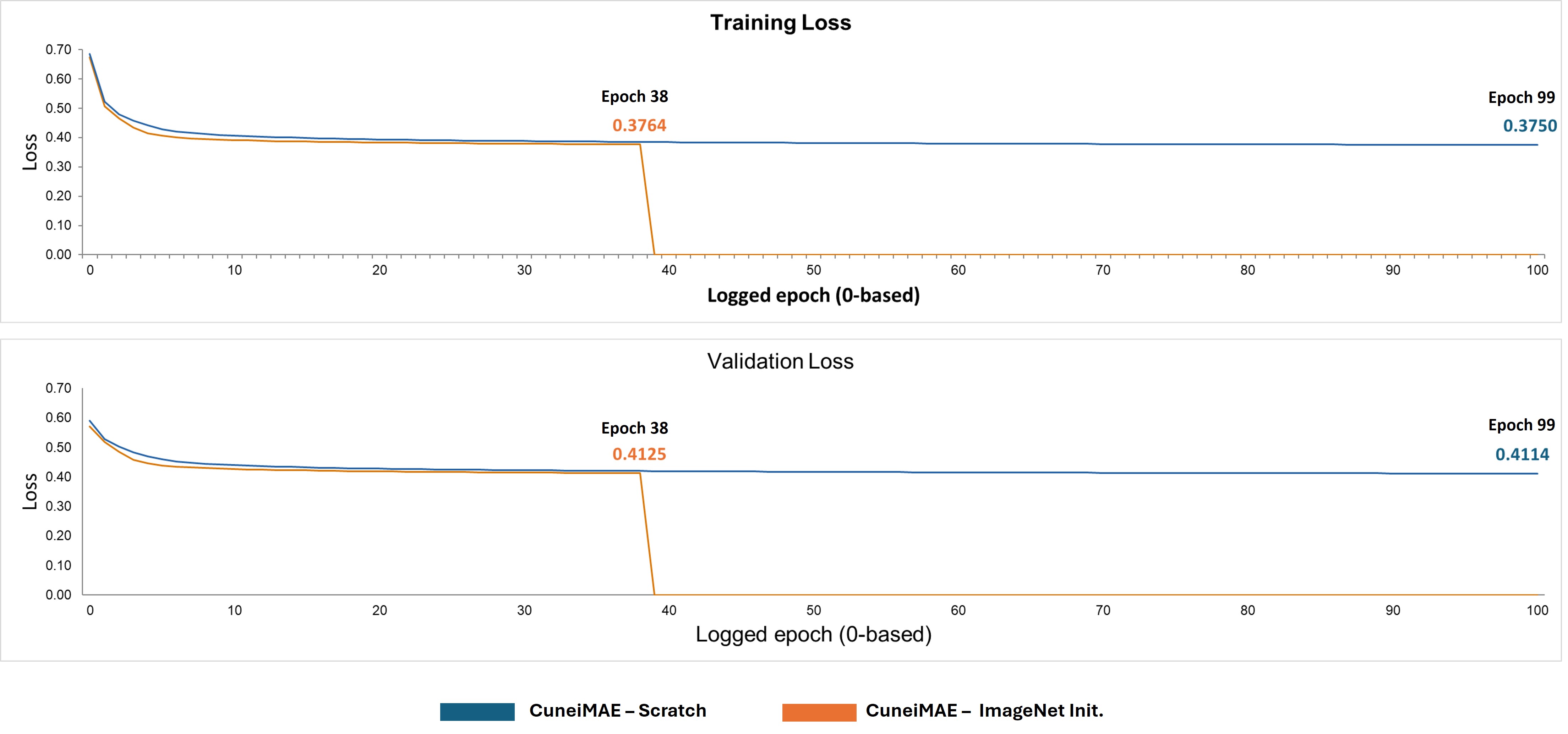}
    \caption{{Training and validation reconstruction losses for scratch- and ImageNet-initialized CuneiMAE. Curves end at their final logged epochs: index 99 (100 completed) and index 38 (39 completed), respectively.}}
    \label{fig:cuneimae_pretraining_loss}
\end{figure*}

{Figure~\ref{fig:cuneimae_pretraining_loss} shows the training and validation reconstruction losses during pretraining. Both decrease over time, with larger improvements in the early epochs and more gradual improvements later. The selected checkpoint, at epoch index 99 after 100 completed epochs, has a validation reconstruction loss of 0.4114 and is used for downstream evaluation.}

\section{Qualitative $k$-NN Retrieval}
\label{supp:qualitative_knn}

\begin{figure*}[t]
    \centering
    \includegraphics[width=\textwidth]{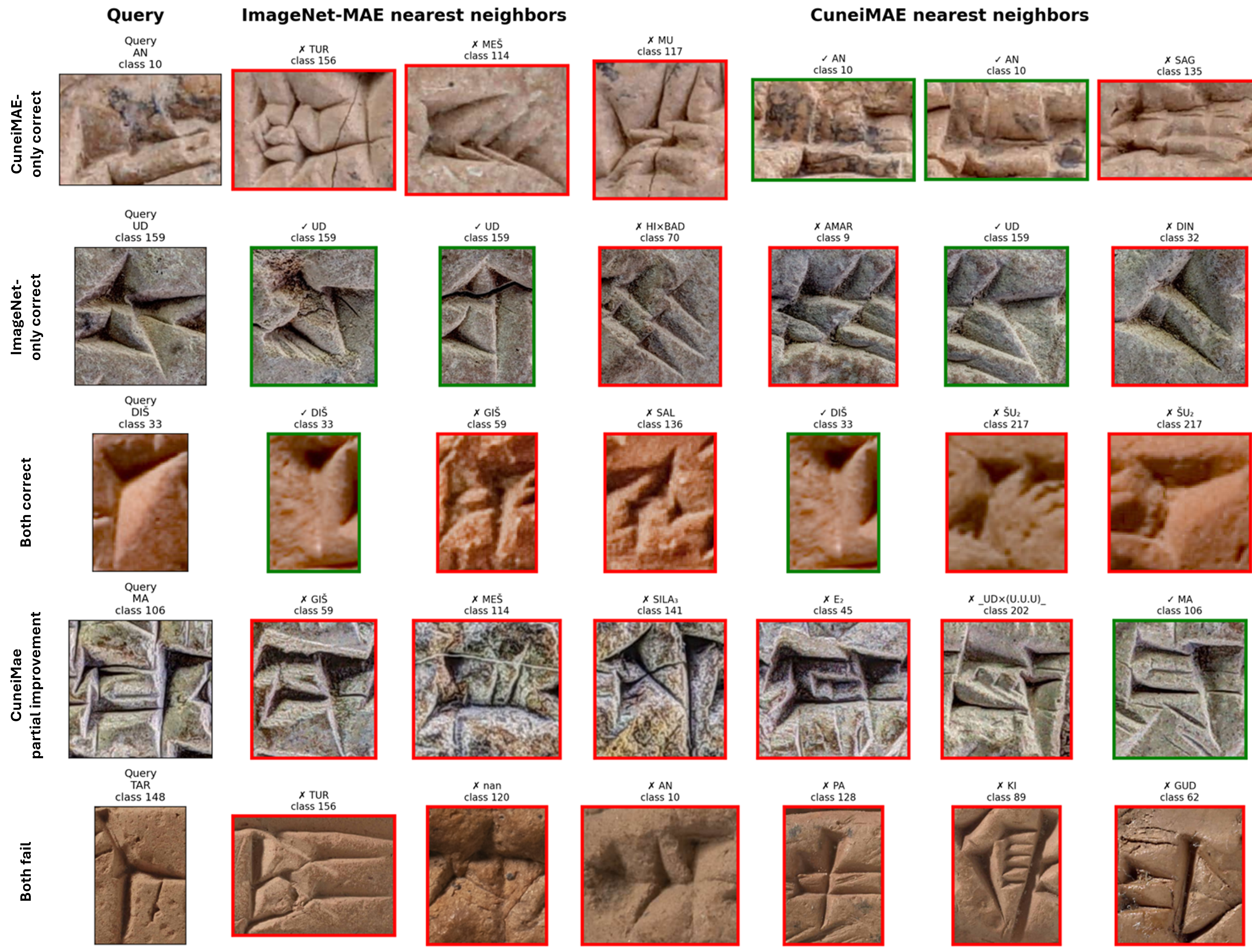}
    \caption{Qualitative comparison of frozen nearest-neighbor retrieval using cosine similarity. Green borders indicate correct-class neighbors and red borders indicate mismatches.}
    \label{fig:supp_qualitative_knn}
\end{figure*}

To complement the quantitative $k$-NN evaluation reported in the main paper, Fig.~\ref{fig:supp_qualitative_knn} shows representative top-3 nearest-neighbor retrieval results using frozen, $L_2$-normalized embeddings and cosine similarity. {Query and gallery images are resized directly to $224\times224$ using bicubic interpolation and normalized using ImageNet statistics, without cropping or flipping.} Queries are drawn from the held-out test split, while neighbors are retrieved from the training gallery. The examples include cases where only ImageNet-MAE succeeds, only CuneiMAE succeeds, both succeed, both fail, and CuneiMAE shows partial improvement.

\section{External CSDD-Derived Evaluation Details}
\label{supp:protosnap_details}

This section provides additional implementation details for the external CSDD-derived experiments described in the main paper.

Overall accuracy denotes the fraction of correctly classified test crops. Rare classes are defined from the original real training split as classes containing fewer than 50 real training examples. Synthetic samples do not alter rare-class membership. Rare accuracy is computed over test samples belonging to these classes. Balanced accuracy is the macro-average of per-class recall, and rare balanced accuracy is the corresponding macro-average over rare classes only.

\subsection{Synthetic-Augmentation Fine-Tuning Protocol}
\label{supp:synthetic_augmentation}
For SD, Stage~1 includes 8,450 synthetic samples covering 169 mapped classes, whereas CN includes 7,800 synthetic samples covering 156 eligible classes. In both cases, 50 synthetic samples are generated per covered class. The complete data counts and training schedule are summarized in Table~\ref{tab:supp_synthetic_protocol}.

The validation set contains 919 real crops and is evaluated after every epoch. The checkpoint achieving the highest validation Top-1 accuracy across both stages is retained. The held-out test set contains 3,442 real crops and is evaluated once after checkpoint selection; it is not used for model selection or hyperparameter tuning.

\begin{table}[h]
\centering
\caption{Two-stage synthetic-augmentation protocol on the reconstructed
187-class CSDD-derived benchmark.}
\label{tab:supp_synthetic_protocol}
\small
\setlength{\tabcolsep}{3.5pt}
\begin{tabular}{lcccc}
\toprule
\textbf{Stage} &
\textbf{Data} &
\textbf{Synthetic} &
\textbf{Total} &
\textbf{Epochs} \\
\midrule
SD Stage 1 & Real + SD & 8,450 & 12,061 & 8 \\
CN Stage 1 & Real + CN & 7,800 & 11,411 & 8 \\
Stage 2 & Real only & -- & 3,611 & 10 \\
Validation & Real only & -- & 919 & Every epoch \\
Final test & Real only & -- & 3,442 & Once \\
\bottomrule
\end{tabular}
\end{table}

\paragraph{ResNet-18.}
The model is initialized from ImageNet-pretrained weights. Both SD and CN experiments use Adam with learning rate $5\times10^{-4}$, batch size 64, and seed 42. {{Training and evaluation images are directly resized to $224\times224$ without cropping, followed by ImageNet normalization.}

\paragraph{CuneiMAE.}
For the CuneiMAE synthetic-augmentation experiments, ViT-B/16 is initialized from the domain-pretrained CuneiMAE checkpoint. Optimization uses Adam with learning rate $1\times10^{-4}$. The physical batch size is 32 with gradient accumulation over two steps, giving an effective batch size of 64. A 5-epoch warm-up is applied independently at the beginning of each training stage. {{Both training and evaluation images are directly resized to $224\times224$ without cropping or horizontal flipping, followed by ImageNet normalization.}

\begin{table}[t]
\centering
\caption{Preprocessing and pretraining-duration ablations on the reconstructed 187-class CSDD-derived test split (\%). {Panel A reports crop-and-flip results. Panel B reports direct-resizing results for CuneiMAE checkpoints pretrained for 10, and 27 epochs.} Synthetic settings use 8 epochs of real+synthetic training followed by 10 epochs of fine-tuning on real data. SD-{\cuneiformA} and CN-{\cuneiformA} denote augmentation with samples generated using the released ProtoSnap diffusion and ControlNet models, respectively. Bal. is Balanced Accuracy. Bold indicates the best result within each model and setting group.}
\label{tab:external_sign_classification_ablation}
\small
\setlength{\tabcolsep}{4pt}
\renewcommand{\arraystretch}{1.05}
\begin{tabular}{lcccc}
\toprule
\textbf{Method / Setting} & \textbf{Acc.} & \textbf{Rare Acc.} & \textbf{Bal.} & \textbf{Rare Bal.} \\
\midrule
\multicolumn{5}{l}{\textit{{Panel A: Crop-and-flip preprocessing}}} \\
\multicolumn{5}{l}{\textit{ResNet-18}} \\
\hspace{0.5em} CSDD & 33.99 & 20.41 & 13.53 & 9.31 \\
\hspace{0.5em} +SD-{\cuneiformA} & 39.19 & $\mathbf{30.16}$ & $\mathbf{20.25}$ & $\mathbf{17.18}$ \\
\hspace{0.5em} +CN-{\cuneiformA} & $\mathbf{42.24}$ & 28.52 & 19.51 & 15.26 \\
\multicolumn{5}{l}{\textit{CuneiMAE {(pretrained 10 epochs)}}} \\
\hspace{0.5em} CSDD & 40.91 & 22.35 & 13.76 & 8.27 \\
\hspace{0.5em} +SD-{\cuneiformA} & $\mathbf{44.92}$ & $\mathbf{31.11}$ & $\mathbf{22.63}$ & $\mathbf{18.39}$ \\
\hspace{0.5em} +CN-{\cuneiformA} & 44.22 & 29.57 & 19.71 & 15.32 \\
\midrule
\multicolumn{5}{l}{\textit{{Panel B: Direct resizing (no crop)}}} \\
\multicolumn{5}{l}{\textit{{ResNet-18}}} \\
\hspace{0.5em} {CSDD} & {58.72} & {46.14} & {37.08} & {32.90} \\
\hspace{0.5em} {+SD-{\cuneiformA}} & {62.99} & \textbf{{52.66}} & \textbf{{40.86}} & \textbf{{37.04}} \\
\hspace{0.5em} {+CN-{\cuneiformA}} & \textbf{{63.80}} & {52.41} & {40.54} & {36.48} \\
\multicolumn{5}{l}{\textit{{CuneiMAE {(pretrained 10 epochs)}}}} \\
\hspace{0.5em} {CSDD} & {58.16} & {46.34} & {34.28} & {30.18} \\
\hspace{0.5em} {+SD-{\cuneiformA}} & \textbf{{62.14}} & \textbf{{51.86}} & \textbf{{40.40}} & \textbf{{36.70}} \\
\hspace{0.5em} {+CN-{\cuneiformA}} & {59.94} & {48.68} & {38.19} & {34.42} \\
\multicolumn{5}{l}{\textit{{CuneiMAE (pretrained 27 epochs)}}} \\
\hspace{0.5em} {CSDD} & {62.67} & {51.27} & {38.63} & {34.50} \\
\hspace{0.5em} {+SD-{\cuneiformA}} & \textbf{{65.33}} & \textbf{{55.89}} & \textbf{{45.82}} & \textbf{{42.52}} \\
\hspace{0.5em} {+CN-{\cuneiformA}} & {63.95} & {53.51} & {43.70} & {40.17} \\
\bottomrule
\end{tabular}
\end{table}

\subsection{Effect of Input Preprocessing}
\label{sec:supp_preprocessing}

{We additionally evaluate an alternative preprocessing configuration under the ProtoSnap-style staged-training protocol. Training images are resized to 256 pixels along the shorter side, followed by random resized cropping to $224\times224$ and horizontal flipping with probability 0.5. Evaluation images are resized to 256 pixels along the shorter side and center-cropped to $224\times224$. ImageNet normalization is applied in both cases.}

{As shown in {Table~\ref{tab:external_sign_classification_ablation}}, crop-and-flip preprocessing substantially reduces classification performance compared with direct resizing {(Panel B of the same table)} {for all settings directly compared across Panels A and B}. For the real-only ResNet-18 baseline, accuracy decreases from 58.72\% to 33.99\%, while {for the real-only CuneiMAE model pretrained for 10 epochs, it decreases} from 58.16\% to 40.91\%. {The 27-epoch CuneiMAE results are shown only for direct resizing and are not part of the preprocessing comparison.}} 

\subsection{External Full Fine-Tuning}

{All models are trained for 90 epochs on real CSDD-derived data without synthetic augmentation. The experiments follow the preprocessing settings described in Sec.~\ref{supp:synthetic_augmentation}, with direct resizing to $224\times224$ and ImageNet normalization, without cropping or flipping. ResNet-18 follows the optimization configuration described above, while the transformer models follow the CuneiMAE fine-tuning configuration.}

\section{Detailed Structure- and Part-Guided MAE Variants}
\label{supp:variants}

This section provides the full implementation details for the structure- and part-guided pretraining variants summarized in the main paper.

\subsection{Edge-Constrained Masking}

For each input crop, we invert channel normalization, clamp pixel values to $[0,1]$, convert the image to grayscale, and compute Sobel-gradient magnitude:
\begin{equation}
\begin{aligned}
G_x &= S_x(x_g), \qquad
G_y = S_y(x_g), \\
E &= \sqrt{G_x^2 + G_y^2 + \epsilon},
\end{aligned}
\end{equation}
where $x_g$ is the grayscale image, $S_x$ and $S_y$ are horizontal and vertical Sobel operators, and $\epsilon$ is a small numerical-stability constant. Reflection padding is used to avoid artificial responses at crop boundaries.

Average pooling over each $16\times16$ patch window yields a patch-level edge-salience score $s_i$. To prevent crop-border artifacts from influencing edge-prioritized sampling, outer patch-grid scores are set to $-\infty$ during candidate selection; these patches remain eligible for subsequent random masking.

For both Edge40 and Edge75, the top 40\% of patches by salience form the candidate set $\mathcal{C}_{\mathrm{edge}}$. We then sample 40\% of these candidates without replacement to obtain the edge-prioritized mask set:
\begin{equation}
|\mathcal{C}_{\mathrm{edge}}|
=
\operatorname{round}(0.40L),
\qquad
|\mathcal{M}_{\mathrm{edge}}|
=
\operatorname{round}(0.40|\mathcal{C}_{\mathrm{edge}}|).
\end{equation}

Edge40 masks only $\mathcal{M}_{\mathrm{edge}}$. Edge75 additionally samples uniformly from the remaining patches until the total masking ratio reaches 75\%:
\begin{equation}
\mathcal{M}_{75}
=
\mathcal{M}_{\mathrm{edge}}
\cup
\mathcal{M}_{\mathrm{rand}},
\qquad
|\mathcal{M}_{75}|=\operatorname{round}(0.75L).
\end{equation}

For the ViT-B/16 patch grid ($L=196$), the candidate set contains 78 patches and the edge-prioritized set contains 31 patches. Edge40 therefore masks 15.8\% of patches. Edge75 adds 116 randomly sampled patches, giving 147 masked and 49 visible patches.

\subsection{Edge-Weighted Reconstruction}

Patch-level Sobel salience is computed as described above, except that border patches are assigned zero salience. Scores are normalized independently for each image:
\begin{equation}
\tilde{s}_i =
\frac{s_i-\min_j s_j}
{\max_j s_j-\min_j s_j+\epsilon}.
\end{equation}
The resulting normalized scores are used in the weighted reconstruction objective defined in the main paper.

\subsection{Dual RGB-Edge Reconstruction}

The dual-target model retains standard random masking and uses one shared ViT encoder with two separate decoders. The RGB decoder reconstructs normalized RGB patch values,
$\hat{x}_i\in\mathbb{R}^{P^2\cdot3}$, whereas the edge decoder reconstructs single-channel Sobel-gradient targets,
$\hat{E}_i\in\mathbb{R}^{P^2}$. Edge targets are computed using reflection padding.

Both decoders are optimized over the same masked-patch set $\mathcal{M}$. The Sobel reconstruction term used in the combined objective defined in the main paper is
\begin{equation}
\mathcal{L}_{\mathrm{sobel}}=
\frac{1}{|\mathcal{M}|}
\sum_{i\in\mathcal{M}}
\lVert E_i-\hat{E}_i\rVert_2^2.
\end{equation}

During downstream evaluation, both decoders are discarded and only the pretrained encoder is retained.

\subsection{Part-Bottleneck MAE}

The part-bottleneck variant introduces iterative Slot Attention between the encoder and decoder. Visible encoder tokens, excluding the [CLS] token, are denoted by
$z_{\mathrm{vis}}\in\mathbb{R}^{L_{\mathrm{vis}}\times D}$. A set of $K=8$ task-agnostic slots competes for visible tokens over $T=3$ iterations. {During pretraining, the slots are initialized by sampling from a Gaussian distribution with learnable mean and log-standard-deviation parameters.} At each iteration, slot-to-token attention is computed as
\begin{equation}
\begin{aligned}
A &=
\operatorname{Softmax}_{\mathrm{slots}}
\left(
\frac{q(S)k(z_{\mathrm{vis}})^{\top}}
{\sqrt{D}}
\right),\\
{\widetilde{A}_{kn}}
&{=
\frac{A_{kn}+\epsilon}
{\sum_{n'}(A_{kn'}+\epsilon)}},\\
U &= {\widetilde{A}}\,v(z_{\mathrm{vis}}).
\end{aligned}
\end{equation}
where the softmax is applied over the slot dimension {and the resulting attention weights are normalized across input tokens for each slot}. Slots are updated using a gated recurrent unit:
\begin{equation}
S^{(t)}=
\operatorname{GRU}\!\left(U^{(t)},S^{(t-1)}\right).
\end{equation}
{Each GRU update is followed by a residual MLP applied to layer-normalized slot representations.}

After the final iteration, positional query tokens for all $L$ patch locations cross-attend to the converged slots before decoding. The decoder therefore cannot directly access encoder patch tokens, forcing reconstruction information through the compact slot bottleneck. {Since the reconstruction objective operates on slot representations derived from encoder patch tokens rather than directly on the [CLS] token, we use mean-pooled encoder patch tokens for frozen linear probing of this variant.}

\end{document}